\documentclass{article}
\usepackage{ijcai23}

\usepackage{times}
\usepackage{soul}
\usepackage{url}
\usepackage[hidelinks]{hyperref}
\usepackage[utf8]{inputenc}
\usepackage[small]{caption}
\usepackage{graphicx}
\usepackage{amsmath}
\usepackage{amsthm}
\usepackage{booktabs}
\usepackage{algorithm}
\usepackage{algorithmic}
\usepackage{subcaption}
\usepackage[switch]{lineno}
\DeclareUnicodeCharacter{2061}{}
\usepackage{siunitx}
\usepackage{caption}

\newcommand\blfootnote[1]{%
  \begingroup
  \renewcommand\thefootnote{}\footnote{#1}%
  \addtocounter{footnote}{-1}%
  \endgroup
}

\title{Using Unsupervised and Supervised Learning and Digital Twin for Deep Convective Ice Storm Classification}

\author{
Jason Swope$^1$
\and
Steve Chien$^1$\and
Emily Dunkel$^1$\and
Xavier Bosch-Lluis$^1$\and
Qing Yue$^1$\And
William Deal$^2$
\affiliations
$^1$Jet Propulsion Laboratory, California Institute of Technology\\
$^2$Northrop Grumman\\
\emails
jason.swope@jpl.nasa.gov
}

\begin{document}

\maketitle
\blfootnote{© 2023. All rights reserved.}
\begin{abstract}
Smart Ice Cloud Sensing (SMICES) is a small-sat concept in which a primary radar intelligently targets ice storms based on information collected by a lookahead radiometer. Critical to the intelligent targeting is accurate identification of storm/cloud types from eight bands of radiance collected by the radiometer. The cloud types of interest are: clear sky, thin cirrus, cirrus, rainy anvil, and convection core. 
 
 We describe multi-step use of Machine Learning and Digital Twin of the Earth's atmosphere to derive such a classifier.  First, a digital twin of Earth's atmosphere called a Weather Research Forecast (WRF) is used generate simulated lookahead radiometer data as well as deeper "science" hidden variables.  The datasets simulate a tropical region over the Caribbean and a non-tropical region over the Atlantic coast of the United States. A K-means clustering over the scientific hidden variables was utilized by human experts to generate an automatic labelling of the data - mapping each physical data point to cloud types by scientists informed by mean/centroids of hidden variables of the clusters. Next, classifiers were trained with the inputs of the simulated radiometer data and its corresponding label. The classifiers of a random decision forest (RDF), support vector machine (SVM), Gaussian naïve bayes, feed forward artificial neural network (ANN), and a convolutional neural network (CNN) were trained. Over the tropical dataset, the best performing classifier was able to identify non-storm and storm clouds with over 80\% accuracy in each class for a held-out test set. Over the non-tropical dataset, the best performing classifier was able to classify non-storm clouds with over 90\% accuracy and storm clouds with over 40\% accuracy.  Additionally both sets of classifiers were shown to be resilient to instrument noise.
\end{abstract}

\section{Introduction}
High altitude ice clouds, covering more than 50\% of the Earth’s surface are often produced from high-impact deep convection events \cite{luo2004}, and are strong modulators of Earth’s weather and climate \cite{stephens2005}\cite{bony2006}. High altitude ice clouds play a significant role in the Earth’s energy balance and hydrologic cycle through their effects on radiative feedback and precipitation, and are therefore crucial for life on Earth.  The SMICES mission is designed to dramatically increase our knowledge of these deep convective ice storms.  

Most satellites blindly image nadir (directly below the orbiting spacecraft) without any knowledge of the phenomena they are observing. Even though global cloud coverage is roughly $2/3$ of the Earth \cite{king2013}, deep convective storms are far rarer than all clouds, representing only a fraction of a percent of all data.  Additionally, radar used to study ice storms has a small footprint of $4kmx4km$. This means that if the radar were to blindly image nadir, very few storms would be captured. Additionally, because radar uses a lot of energy, SMICES will only be able to capture data for about 20\% of the time.   These constraints lead to the need for active targeting, which will allow the mission to maximize its scientific return by selecting which clouds are analyzed.

This paper focuses on the classification system of the SMICES mission. We describe the use of unsupervised machine learning, digital twin (WRF), and supervised machine learning to address the challenge of developing a radiometer-based classifier for deep convective ice storms.

The remainder of this paper is organized as follows. In Sections 3 we provide context on the problem and an overview of the classifiers and data that will be used. Section 4 describes the tropical and non-tropical datasets and how an automatic labelling system was established. Section 5 and 6 explain the different classifiers utilized and how they were set up for this experiment. Section 7 reviews the results of each classifier on both datasets. Section 8 explores the impact of noise on the accuracy of the classifiers. Section 9 and 10 discusses the results of the classifiers in the context of overall performance and outlines future work. Finally, in section 11 we summarize our conclusions.

\section{Related Work}

There are examples of work on cloud detection in other use cases such as cloud avoidance. On the EMIT mission cloud detection is used on the hyperspectral data to screen out covered sections of images \cite{Oaida}. Other uses of cloud screening can be found in onboard flood and cryospheric classification \cite{IpFloodDetection}\cite{DOGGETT2006447}. Further work on screening clouds out of data data has been pursued to prepare for the unprecedented data volumes that will come from future missions \cite{Thompson}. 

In our work, we use supervised learning for storm classification at the pixel level. We focus on storm classification from satellite data instead of cloud detection, and we use simulated radiometer data from a satellite. In addition, we use a digital twin in combination with sparse labeling and unsupervised clustering to facilitate label generation.

Prior work has explored the use of unsupervised learning to cluster cloud data. Clustering has been explored using three-dimensional histograms applied to multi-spectral satellite imagery, using the visible, IR, and water-vapor channels \cite{desbois1982automatic}. Later work improved upon this by introducing textural parameters and processing larger datasets at different times \cite{seze1987cloud}. Another work attempted to reproduce the class clusters using Probabilistic Self-organizing Maps \cite{AMBROISE200047}.

Our clustering approach differs from the above work, in that we use simulated science parameters from a digital twin to do clustering, as opposed to using features derived from imagery. To the best of our knowledge, there is currently no other work with digital twins being used for the purpose of facilitating label generation in cloud imagery.

\section{Background}
The SMICES classification problem is to correctly classify different cloud types utilizing the information that the on-board radiometer collects. The classifier needs to be able to correctly identify the cloud as one of five cloud types: clear sky, thin cirrus, cirrus, rainy anvil, and convection core. The scientifically highest value cloud is the convection core followed by the rainy anvil cloud. With respect to the SMICES targeting algorithms, only the rainy anvil and convection core clouds are actively targeted. Therefore, it is extremely important for the classifiers to be able to distinguish between the rainy anvil, the convection core and the three non-storm cloud types. An accurate classification of these cloud types has been shown to enable a gain of capturing convection core clouds by a factor of 24 and rainy anvil clouds by a factor of 2 \cite{swope_smices_iwpss_2021}.

% \begin{figure}[h]
%   \centering
%   \includegraphics[width=\linewidth]{figures/cloud_types.png}
%   \caption{Cloud types shown in order of increasing interest}
%   \label{fig:cloud_types}
%   \Description{Cloud types}
% \end{figure}

We first use clustering and human experts to label the digital twin / WRF data.  The digital twin datasets include three scientific variables of ice water path (IWP), median particle size (\SI{}{\micro\meter}), and median cloud top height (m), as well as the eight bands of radiance previously mentioned. These three scientific variables are used to automatically generate labels for the data with the help of scientists as explained in section 4.2. This avoids the need to manually label the data. 

We then use these labels with the digital twin (simulated) radiometer data to train the classifiers.  During testing, the classifiers will only have access to the radiance values since those are the only data available in an operational setting. 

A set of classifiers were trained and tested on two separate regional datasets: a tropical dataset over the Caribbean, and a non-tropical dataset of the Atlantic Coast of the United States. The physics of a given cloud type can differ depending on whether it is in a tropical or non-tropical region. For example, there is ice in the deep convective core in non-tropical regions but not in tropical regions, and this affects the radiance values. Thus, we train separate classifiers for each dataset. These classifiers include a random decision forest (RDF), support vector machine (SVM), Gaussian Naïve Bayes, feed forward artificial neural network (ANN), and convolutional neural network (CNN). In an operational setting, we would select the model that corresponds to the geographical region the satellite is over.

\begin{figure*}[ht]
  \centering
  \includegraphics[height = 5.5cm]{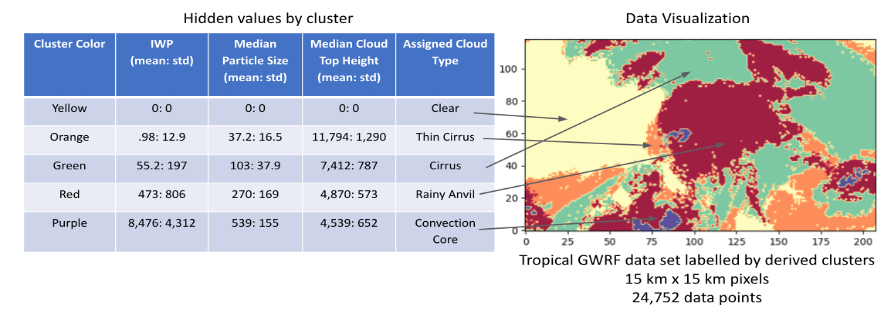}
  \caption{Cluster values mapped to their labels on the tropical dataset}
  \label{fig:clusters}
\end{figure*}

\section{Data}
\subsection{Datasets}
Two regional datasets were used in this work: a tropical and non-tropical dataset. Both datasets were created through the Global Weather Research and Forecasting (GWRF) model \cite{Skamarock2019ADO}. The GWRF is a state-of-the-art physics-based weather model. It is used to create computationally expensive datasets that we use as a digital twin to real climate data. In our case the model generated the brightness temperatures for different cloud types along the bands of Tb250+0.0, Tb310+2.5, Tb380-0.8, Tb380-1.8, Tb380-3.3, Tb380-6.2, Tb380-9.5, and Tb670+0.0, as well as the scientific variables of ice water path, median particle size, and median cloud top height.

The tropical dataset is located in the Caribbean. This dataset contains 13 images that are 119x208 pixels with a pixel size of 15km for a spatial extent of 1,785km x 3,120km. Each image is a snapshot of the same area in one-hour intervals.

The non-tropical dataset is located in the Atlantic Coast of the United States. The data contains 29 image cutouts that are 1998x270 pixels with a pixel size of 1.33km for a spatial extent of 2,657km x 359km. The dataset combines to form three images over the same area in 12-hour time intervals. Each larger image is constructed of 10 image cutouts stacked vertically. The total size of a full image is 1998x2700 pixels with a spatial extent of 2,657km x 3,591km. The last cutout of the third time interval was incomplete and therefore left out of this study. Therefore, we only have 29 image cutouts.

The same cloud types in the tropical vs. non-tropical datasets do not necessarily correspond to similar scientific features or radiance values. Because of this, as well as the large difference in resolution (pixel size of 15km vs. 1km), we treat the tropical and non-tropical datasets independently in this work.

\subsection{Data Labelling Using Clustering and Digital Twin}

Manually labelling our data was not a feasible option given the extremely large number of pixels, and the difficulty in identifying the cloud type based only on radiance values. We used a digital twin of the radiometer to solve this problem. The digital twin helps in two ways. It gives us access to scientific variables that are not directly measurable in nature and also allows us to generate a large amount of storm data for training and evaluation.

The scientific variables of ice water path, median particle size, and median cloud top height are significant because they give us a way to map a pixel’s values to a specific cloud type. Unlike the radiance values, the scientists are able to map these scientific variables to the five cloud types they identified. While this solves the mapping problem, an automated method needed to be developed due to the large number of pixels 

The approach to automate the labelling is as follows. First, we cluster the data into representative clusters based on the scientific variables. Then we can map each cluster center to the cloud type that it corresponds to. Each cluster’s mapping serves as the label for every pixel within that cluster. 

K-means was utilized as our clustering method. This approach requires us to identify the proper number of clusters to accurately represent our dataset before mapping the clusters to the specific cloud types. A failure to accurately represent our data will cause a poor classification of pixels. The proper number of clusters is determined by analyzing if the number of clusters selected maximizes the within cluster coherence and the between cluster separation. A silhouette score was used to calculate the effectiveness of the number of clusters over the data.

A silhouette score, which ranges from -1 to 1, gives a measure of the cluster's fit. A score of 1 means the clusters are well distinguished, a score of 0 means the clusters are indifferent, and a score of -1 means that the clusters are incorrect. The equation used to obtain this score is: \(((b-a)/max⁡(a,b))\) where a represents the intra-cluster distance (average distance between each point within a cluster) and b represents the average nearest cluster distance (average distance between the instances of the next closest cluster). Clustering was performed separately for the tropical and non-tropical datasets.

\begin{figure}[h]
     \centering
     \includegraphics[height=5cm]{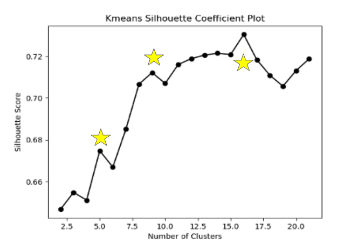}
     \caption{Silhouette scores over the tropical dataset}
     \label{fig:tropical_silhouette}
\end{figure}
 
There are three local maximums at 5, 9, and 16 clusters in the tropical dataset. It is possible to map multiple clusters from K-means to one cloud type if multiple clusters have similar scientific values. The difference between the silhouette scores of the local maximum at five clusters (.675) is only .06 away from the silhouette score at 16 clusters (.735), which is an insignificant increase. We chose to use five clusters due to the high silhouette score and the easier mapping to the five alluded cloud types.

Silhouette scores were also calculated for a random sample taken from the non-tropical dataset. The scores for all of the clustering values were over .97, which suggests a very good clustering for the dataset. We chose to continue to use a 5 cluster K-means for the non-tropical dataset for its high silhouette score and easy mapping to the original cloud labels.

The scientists assigned each cluster to its respective label based on the mean and standard deviation of its centroid. Overall, the clusters gradually increased in IWP and median particle size while decreasing in median cloud top height. The only exception to this rule was the cluster with all values set to 0, which we assign to the clear sky class. This relationship between the clusters was used by the scientists to identify the labels. In general, a higher IWP and median particle size while having lower median cloud top height correlates to a stronger storm cloud. These mappings for the tropical dataset are demonstrated in figure \ref{fig:clusters}.

\section{Classifiers}
The ground truth values for each cloud class are taken to be the labels assigned through the clustering method described above. Now that we have labels, we can train the classifiers to predict the cloud class using only data available in orbit: the radiometer data. During training, the classifiers see both the radiometer data and the class labels, and at test time, see only the radiometer data. The classifiers operate at the single pixel level; given a radiometer measurement at a given pixel, what is the cloud class at that pixel. Thus, our current classifiers do not take into account neighboring pixel values. Classifier performance was evaluated using 5-fold cross-validation, and testing was performed on a separate held-out test set. 

For this study we explored the following classifiers: random decision forest (RDF), support vector machine (SVM), Gaussian Naïve Bayesian, a feed forward artificial neural network (ANN), and a convolutional neural network (CNN). 

An important factor in the classifier configuration is how balanced the different classes within the datasets are. The classes are heavily skewed away from the most important cloud type, the convection core, in both the tropical and non-tropical datasets.

\begin{table}[h]
 \centering
     \begin{subtable}[b]{\linewidth}
         \centering
         \includegraphics[width=\textwidth]{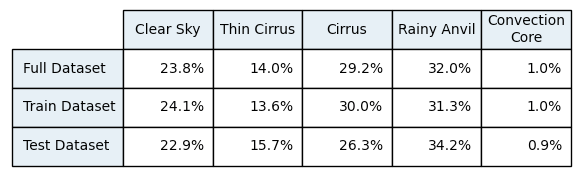}
         \caption{Distributions over the Tropical Dataset}
         \label{tab:trop_class_dist}
     \end{subtable}
     \vfill
     \begin{subtable}[b]{\linewidth}
         \centering
         \includegraphics[width=\textwidth]{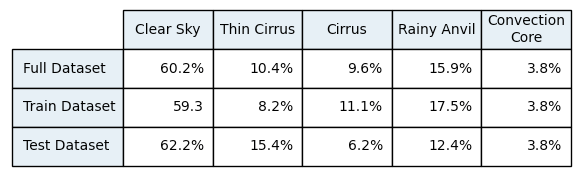}
         \caption{Distributions over the Non-Tropical Dataset}
         \label{tab:non_trop_class_dist}
     \end{subtable}
        \caption{Cloud type distribution over the tropical and non-tropical datasets and their respective train and test datasets}
        \label{tab:class_dist}
\end{table}

Table \ref{tab:class_dist} displays the distribution of cloud type in the tropical and non-tropical dataset. The two most important classes, convection core and rainy anvil, only make up around 32\% of the dataset. Therefore, if we only classified every pixel as clear sky, thin cirrus, or cirrus we would be able to attain a 68\% accuracy, although we would miss the important classes. The class distribution of the non-tropical dataset (table \ref{tab:class_dist}) is significantly more skewed towards clear than the tropical dataset. About 80\% of this dataset is non-storm clouds. To help the classifiers overcome these unbalanced datasets, the random forest and support vector machine classifiers are trained with weights adjusted for the class imbalance. When it is noted that the weights are equalized, we are stating that the weights have been rebalanced so that every class is weighted equally.

\section{Experimental Design}

Separate classifiers were trained for tropical and non-tropical regions, using the data corresponding to that region. For the tropical dataset the first ten images were used as a training and validation set. The remaining three images then served as the test set. The test and validation set for the non-tropical dataset was created from the first two time steps. These two timesteps are comprised of the first 20 image cutouts in the dataset. The remaining 9 image cutouts that makeup the final image was used as the test set. The distribution of the cloud classes in each training and test dataset is shown in table \ref{tab:class_dist}. We did not build classifiers using a combined dataset due to the different pixel sizes of each dataset and physical differences found in tropical and non-tropical clouds.

The classifiers are evaluated on their performance by analyzing their accuracy over the three cloud classes of non-storm (clear, thin cirrus, and cirrus), rainy anvil, and convection core. The classifiers were trained in the following method:
\begin{enumerate}
    \item Trained on the original five-class labelling created through K-means. These classifiers output a five-class labelling of the clouds which is converted into the three-class accuracy\footnote{We also explored training the data on three class labeled data, however this produced similar results. This paper will focus on the 5 class results.}
\end{enumerate}

The performance of each classifier will be presented over the three-class problem of identifying non-storm clouds (clear, thin cirrus, or cirrus), rainy anvil clouds, and convection core clouds. The simplification from five to three classes is due to SMICES only actively targeting rainy anvil and convection core clouds, with a preference for convection core. Therefore, the meaningful distinctions are between those three classes. However, not all misclassifications are equal. A misclassification of a non-storm cloud as rainy anvil or convection core cloud could trigger the use of radar power for a non-desired target. This would be more costly than a misclassification between a convection core and rainy anvil cloud since the radar would still be collecting scientifically significant data.
The performance will also be discussed in the context to the two class problem of non-storm clouds (clear, thin cirrus, and cirrus), and storm clouds (rainy anvil and convection core). This analysis highlights how many significant misclassifications are being made by each classifier.

\section{Classifier Results}
\subsection{Random Decision Forest (RDF)}
We use scikit-learn's implementation of the Random Forest Classifier in this work \cite{scikit-learn}. We found that 32 trees and a maximum depth of 14 was optimal based on the training and validation sets.

One challenge with the storm datasets is their unbalanced nature. Storms are quite rare in the sky, and the center of storms (convection core clouds) are even rarer. Therefore, the classifier is going to be more prone to classifying clouds as non-storm clouds, since that is the most dominant class. To solve this, the weights have been adjusted in the RDF so that every class is weighted equally.

% Table \ref{tab:rdf_balanced} compares the performance of the RDF classifier over the tropical dataset trained on five-class labelled data when the weights of the RDF are equalized and not equalized by class. Overall, we see that balancing the weights increases our accuracy in the convection core class by 11\% and gives a small increase in the non-storm class. The disadvantage of this method is a 4\% decrease in the accuracy of rainy anvil clouds as well as an increase in the misclassification of both rain anvil and convection core clouds as non-storm. This is a more costly misclassification since they could cause SMICES to expend energy over a non-interesting target. Overall, the performance gained by balancing the class weights is significant because of the increase in the accuracy of the most important cloud type (convection core).  

\begin{table}[h]
    \centering
    \includegraphics[width=8.5cm]{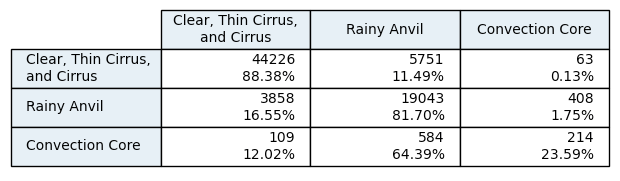}
     \label{tab:rdf_trop_3}
    \caption{Confusion Matrix of the RDF classifier on the tropical dataset, max depth 14, number of trees 32, weights equalized by class, 3 class labeled data}
    \label{tab:rdf_tropical}
\end{table}

Table \ref{tab:rdf_tropical} shows the RDF performance. On the tropical data the classifier was able to accurately classify non-storm clouds and rainy anvil clouds with 88\% and 82\% and accuracy respectively. Even though the classifier failed to accurately identify convection core clouds, the majority of convection core misclassifications were as rainy anvil clouds. When looking at the storm cloud accuracy the classifier was able to identify storm clouds with 84\% accuracy.

\begin{table}[h]
     \centering
     \includegraphics[width=8.5cm]{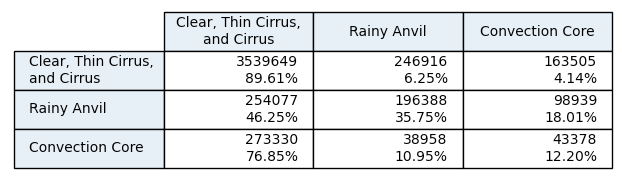}
    \caption{Confusion Matrix of the RDF classifier on the non-tropical dataset, max depth 14, number of trees 32, weights equalized}
    \label{tab:rdf_non_trop}
\end{table}

The non-tropical dataset proved to be significantly more difficult to classify than the tropical dataset. The accuracy for the storm cloud classes (rainy anvil and convection core) decreased by about 2x. The non-storm class accuracy remained relatively unchanged when compared to the tropical dataset. When looking at the non-storm/storm cloud accuracy the RDF achieved 90\% non-storm accuracy and 42\% storm accuracy.

\subsection{Support Vector Machine (SVM)}
The support vector machine used was the linear support vector classifier (linear SVC) from Scikit Learn. This classifier is suggested  if the dataset contains greater than tens of thousands of data points by Scikit Learn. The linear SVC is effectively an support vector machine with a linear kernel.

\begin{table}[h]
     \centering
     \includegraphics[width=8.5cm]{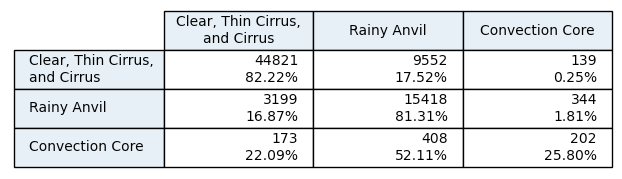}
    \caption{Confusion Matrix of the linear SVC classifier on the tropical dataset, weights balanced}
    \label{tab:svm_trop}
\end{table}

Table \ref{tab:svm_trop} demonstrates the accuracy of the SVM classifier with a linear kernel on the tropical dataset. The classifier accurately classifies the non-storm and rainy anvil classes with 82\% and 81\% accuracy. The SVM failed to accurately classify the convection core class. Similar to the RDF, most of its convection core misclassifications were labeled as rainy anvil. The storm cloud accuracy for this classifier is 68\%.

\begin{table}[h]
     \centering
     \includegraphics[width=8.5cm]{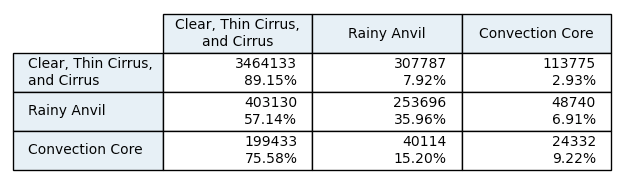}
    \caption{Confusion Matrix of the linear SVC classifier on the non-tropical dataset, weights balanced}
    \label{tab:svm_non_trop}
\end{table}

The SVM classifier performed worse in the storm classes on the non-tropical dataset than it did on the tropical dataset. While the non-storm classification accuracy increased to 89\%, the accuracies in both storm classes decreased by more than half when compared to the tropical dataset performance. The storm class accuracy over the non-tropical dataset 40\%.

\subsection{Gaussian Naïve Bayes}

The Gaussian Naïve Bayes classifier used is from Scikit learn. The likelihood of each feature is assumed to be Gaussian.

\begin{table}[h]
     \centering
     \includegraphics[width=8.5cm]{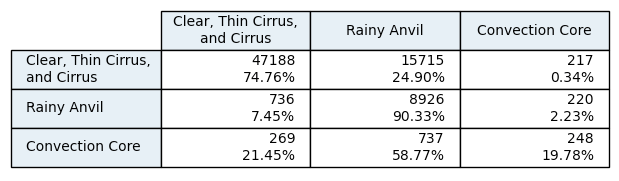}
     \label{tab:gnb_trop_3}
    \caption{Confusion Matrix of the Gaussian Naïve Bayes Classifier on the tropical dataset}
    \label{tab:gnb_trop}
\end{table}

Table \ref{tab:gnb_trop} contains the results of the Gaussian Naïve Bayes classifier when trained on the tropical dataset. It achieved the best classification in the rainy anvil class with 90\%, however struggled with the non-storm and convection core classes compared to the other classifiers. Most of the convection core misclassificaitons were made as rainy anvil classifications. The two class accuracy over the tropical dataset is 75\% for non-storm and 42\% for storm.

\begin{table}[h]
     \centering
     \includegraphics[width=8.5cm]{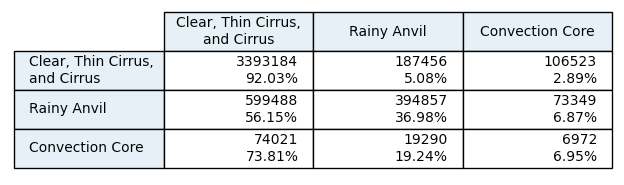}
     \label{tab:gnb_non_trop_3}
    \caption{Confusion Matrix of the Gaussian Naïve Bayes Classifier on the non-tropical dataset}
    \label{tab:gnb_non_trop}
\end{table}

Table \ref{tab:gnb_non_trop} shows the performance of the GNB classifiers over the non-tropical dataset. Similar to the RDF and SVM, the GNB struggled to accurately classify the separate classes. While the non-storm accuracy increased when compared to the tropical dataset, its accuracy in both storm classes decreased by more than half. The two class accuracy over the tropical dataset is 92\% for non-storm and 55\% for storm.

\subsection{Neural Networks}
We also applied the use of simple neural networks using Keras \cite{chollet2015keras}  and TensorFlow \cite{tensorflow2015-whitepaper}.

\subsubsection{Feed Forward Artificial Neural Network (ANN)}

We pass each 8-band pixel into a feed forward ANN with 2 hidden layers, 32 nodes in each hidden layer, a dropout rate of 0.1, and a softmax activation at the final layer. We used ADAM \cite{kingma2017adam} as our optimizer. Model performance was not affected by small changes in these parameters. The model output is a vector with probability of each cloud-type class, and we take the Argmax to get the predicted class.

\begin{table}[h]
     \centering
         \includegraphics[width=8.5cm]{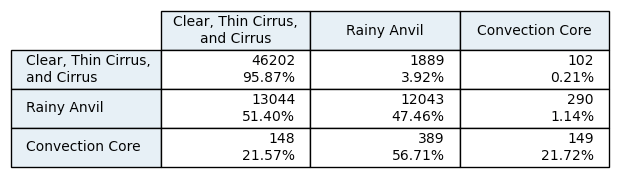}
         \label{tab:ann_trop_3}
        \caption{Confusion Matrix of the ANN on the tropical dataset}
        \label{tab:ann_trop}
\end{table}

The ANN performance over the tropical dataset was relatively poor compared to the past classifiers. While the classifier achieved around 95\% accuracy in the non-storm class, it struggled to identify both the convection core and rainy anvil clouds (table \ref{tab:ann_trop}). Even in the two class problem neither classifier was able to identify storm clouds with 60\% accuracy. 

The ANN was not effective on the non-tropical dataset, and labelled all pixels as the non-storm class during training and validation. This is probably due to the larger skew of the non-tropical dataset towards the non-storm class.

\subsubsection{Convolutional Neural Network (CNN)}

We also test a single-pixel CNN model. Our architecture uses 3 stacked 1-d convolutional layers, where the convolutions are applied over the bands. Due to the small input size, we did not apply pooling. 6, 12, and 24 filters were used for each convolutional layer respectively, and a rectified linear unit (relu) activation was used. We found a convolutional filter shape of 1x3 to be optimal. The last layers of our net included two fully connected layers with 32 nodes, a dropout of 0.1, and a final sofmax activation. ADAM optimizer was used to train the network.

\begin{table}[h]
         \includegraphics[width=8.5cm]{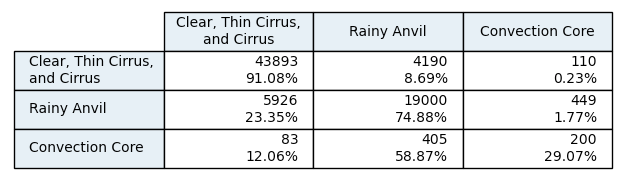}
         \label{tab:cnn_trop_3}
        \caption{Confusion Matrix of the CNN on the tropical dataset}
        \label{tab:cnn_trop}
\end{table}

Table \ref{tab:cnn_trop} show the CNN results on the tropical dataset. The classifier was able to classifiy both non-storm and rainy anvil clouds accurately with over 91\% accuracy in non-storms and 75\% accuracy in rainy anvil. The classifier still struggled with convection core. It was able to achieve a storm cloud accuracy of 83\% which is very close to the RDF performance.  

\begin{table}[h]
         \includegraphics[width=8.5cm]{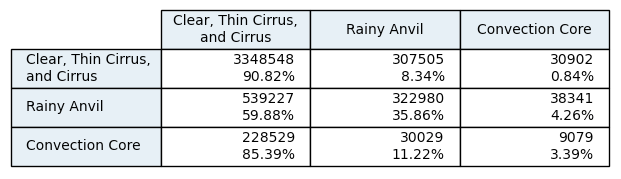}
         \label{tab:cnn_non_trop_3}
        \caption{Confusion Matrix of the CNN on the non-tropical dataset}
        \label{tab:cnn_non_trop}
\end{table}

The CNN experiences a similar outcome to the ANN when trained on the non-tropical dataset. The heavier skew towards non-storm clouds caused the classifier to classify the majority of the test dataset as non-storm. The classifier achieved a 91\% accuracy in non-storm clouds and a 44\% accuracy in storm clouds. For future work, we plan to upsample the minority classes when training our neural networks. 

\section{Noise}
In flight we expect the sensor to generate Gaussian noise that is independent along each radiance band. The Tb380 band is expected to have around 5 kelvin of noise, and the rest of the bands are expected to have 1 kelvin.

To test the impact of the expected noise on the classifiers we generate random values from a Gaussian distribution with a mean of 0 and a standard deviation of the expected noise for each band. These values are then independently added to each band to create a noisy dataset. The impact of noise is determined by the total decrease in accuracy of the three-class RDF classifier when tested on the clean held-out test set, and on the same testset, but with noise applied. 

\begin{enumerate}
    \item Tropical dataset: RDF classifier with a max depth of 14, 32 trees, and weights equalized was trained on the first eight images of the tropical dataset. It was then tested on the remaining five images after expected noise was applied to those data values.
    \item Non-tropical dataset: RDF classifier with a max depth of 14, 32 trees, and weights equalized was trained on the first image of the non-tropical dataset (10 image cutouts). It was then tested on the second image (10 image cutouts) after expected noise was applied noise was applied to those data values.
\end{enumerate}

Over the tropical dataset the expected noise in flight decreased the accuracy of the RDF by 4\%. When the noise was applied to the non-tropical dataset the accuracy of the RDF decreased by 2\%. This demonstrates that the RDF classifier is realatively robust to noise in both datasets

\section{Discussion}

Multiple classifiers were able to accurately distinguish between the two-class problem in the tropical dataset. It is important to understand that even less accurate classifiers can be used to significantly increase yield of the SMICES mission concept.  In Table \ref{tab:trop_targeting_dist} the first row shows the distribution of pixels acquired if sampling pixels at random (e.g. not using any classification). Row two shows the distribution if we sample pixels the RDF classifies as Rainy Anvil. Note that in the chart, the term "Sample Labelled" means what the classifier thinks is the correct class. We still get some non-storm and Convection Core pixels, since the classifier is not 100\% accurate. Row three shows the distribution if we only sample pixels the RDF classifies as Convection Core. Again, we get other classes as well due to inaccuracies in the classifier.  This data shows that sampling a Rainy Anvil or Convection Core classified pixel is far more fruitful scientifically than a random pixel. 

Previous work by Swope \cite{swope_smices_iwpss_2021} has evaluated the improvements in mission return by running mission simulations using WRF datasets in which intelligent targeting attempts to preferentially target areas of Convection Core and Rainy Anvil but is limited by rarity of such pixels and mission energy and pointing constraints. In Table \ref{tab:trop_targeting_dist}, row 4 shows the distribution of pixels we would acquire if the classifier used in the simulation was 100\% accurate. Row 5 shows the distribution of pixels (true labels) we actually acquire, quantifying how classification inaccuracy reduces the impact on return. However, this mode still dramatically outperforms uninformed (random) targeting. This highlights how preferential targeting is able to skew sampled pixels towards storm-cloud classes.

\begin{table}[h]
     \centering
     \begin{subtable}[b]{\linewidth}
         \centering
         \includegraphics[width=\textwidth]{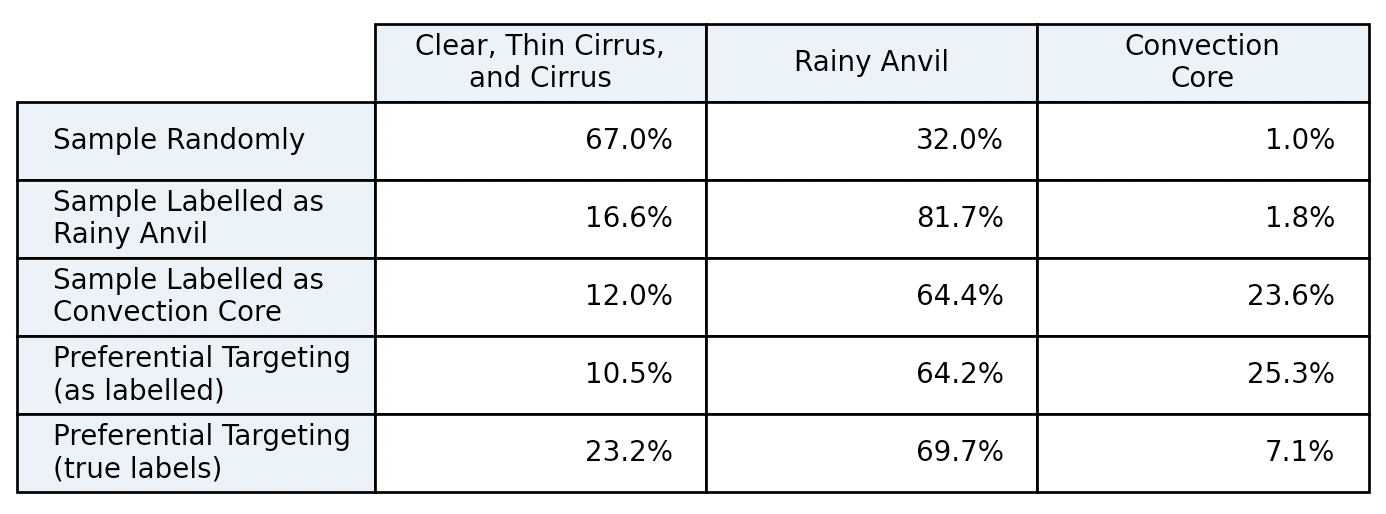}
         \caption{Tropical Data}
         \label{tab:trop_targeting_dist}
     \end{subtable}
     \vfill
     \begin{subtable}[b]{\linewidth}
         \centering
         \includegraphics[width=\textwidth]{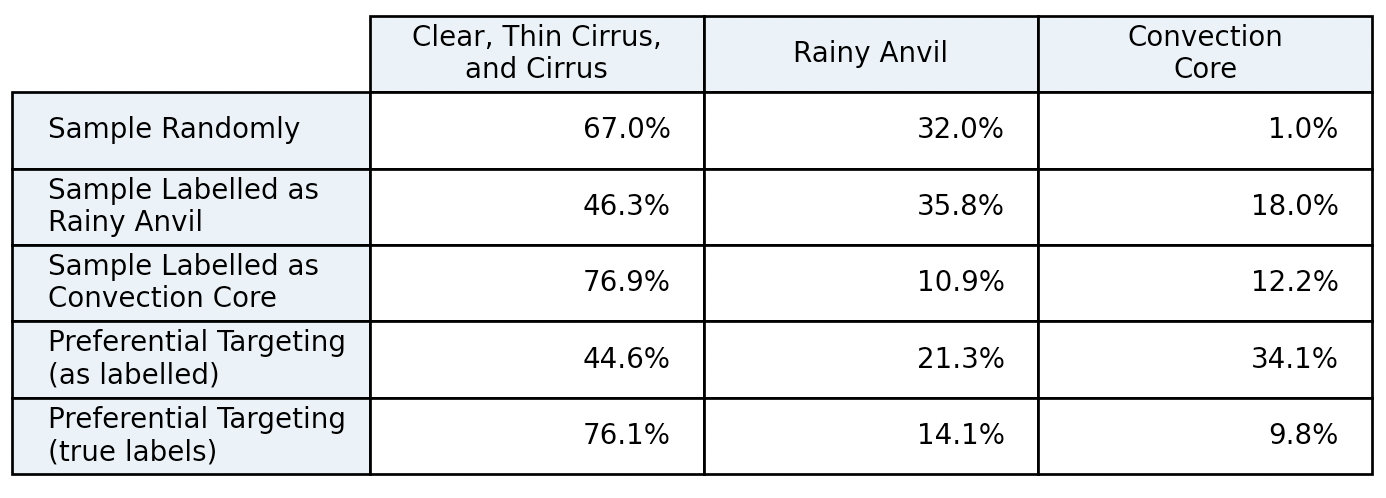}
         \caption{Non-Tropical Data}
         \label{tab:non_trop_targeting_dist}
     \end{subtable}
        \caption{Mission return impact of Intelligent Targeting with Classifier}
        \label{tab:targeting_dist}
\end{table}

When looking at the non-tropical dataset, distinguishing between the two classes of Rainy Anvil and convection core is significantly harder. Again, even with the lower classification accuracy, Table \ref{tab:non_trop_targeting_dist} shows the expected pixels observed sampling from the Rainy Anvil and Convection Core labelled pixels compared to random sampling.  Again, drawing on the mission simulations, the last two rows of table \ref{tab:non_trop_targeting_dist} show dramatically increased yields of Rainy Anvil and Convection Core measurements from intelligent targeting compared to uninformed (random) targeting even with imperfect classification.  

% comment on why the tropical dataset was easier to classify than the non-tropical

\section{Future Work}

We would like to extend to additional regions beyond the Caribbean and Atlantic coast to make our classifiers more robust. A global dataset could be used to explore the possibility of a universal classifier that would work in any region (and season, and other conditions), however variations in atmospheric phenomena in different regions could make this difficult. Swapping between different regional classifiers in flight would be feasible.

The impact of the expected noise should also be analyzed more clearly. Even though the overall accuracy is not strongly impacted, it is important to know if any cloud types are being disproportionately affected by the noise or if is balanced.

Future work on the classifiers will expand beyond single pixel classification and take into account surrounding pixels. This should improve the accuracy because storm phenomena are not randomly distributed across the sky, instead they are clustered close together. Upsampling on the storm clouds in each dataset may also improve the overall performance, for the classifiers that did not have their weights equalized, due to how imbalanced the datasets are.

We intend to test these classifiers on real data from airborne tests of the SMICES radar.

\section{Conclusion}
We have described an effort to develop a classifier of deep convective storms based on radiometer data.  Using a digital twin and K-means clustering we were able to generate labels for data. The results of the classifiers are promising for distinguishing deep convective storms in the tropical dataset. Further work still needs to be done for finer grained storm type discrimination. Additionally, identification of deep convective storms in non tropical data was more challenging.  We also present results indicating that even moderate classification accuracies combined with intelligence instrument targeting are expected to enable significant improvements in mission return. 

\section{Acknowledgements}
The research was carried out at the Jet Propulsion Laboratory, California Institute of Technology, under a contract with the National Aeronautics and Space Administration.

\bibliographystyle{named}
\bibliography{refs}

\end{document}